\documentclass[sigconf,author]{acmart}
\AtBeginDocument{%
  }

\usepackage{array}
\usepackage{tabularray}
\usepackage{algorithm}
\usepackage{algpseudocode}
\usepackage{subcaption}
\usepackage{multirow}

\setcopyright{acmlicensed}
\copyrightyear{2025}
\acmYear{2025}
\acmDOI{XXXXXXX.XXXXXXX}
\acmConference[AIR '25]{Make sure to enter the correct
  conference title from your rights confirmation email}{July 02--05,
  2025}{Jodhpur, India}
\acmISBN{978-1-4503-XXXX-X/2025/07}

\begin{document}

\title{Towards Global Localization Using Multi-Modal Object-Instance Re-Identification}

\author{Aneesh Chavan}
\email{c.aneesh.1203@gmail.com}
\affiliation{%
  \institution{RRC, IIIT Hyderabad}
  \country{India}
}

\author{Vaibhav Agrawal}
\authornote{These authors contributed equally to this research.}
\email{vaibhav.agrawal@research.iiit.ac.in}
\affiliation{%
  \institution{RRC, IIIT Hyderabad}
  \country{India}
}

\author{Vineeth Bhat}
\email{vineeth.bhat@students.iiit.ac.in}
\authornotemark[1]
\affiliation{%
  \institution{RRC, IIIT Hyderabad}
  \country{India}
}

\author{Sarthak Chittawar}
\email{sarthak.chittawar@research.iiit.ac.in}
\authornotemark[1]
\affiliation{%
  \institution{RRC, IIIT Hyderabad}
  \country{India}
}

\author{Siddharth Srivastava}
\email{siddharth.sri89@gmail.com}
\affiliation{%
  \institution{Typeface Inc.}
  \country{India}
}

\author{Chetan Arora}
\email{chetan@cse.iitd.ac.in}
\affiliation{%
  \institution{IIT Delhi}
  \country{India}
}

\author{K Madhava Krishna}
\email{mkrishna@iiit.ac.in}
\affiliation{%
  \institution{RRC, IIIT Hyderabad}
  \country{India}
}




\renewcommand{\thefootnote}{\alph{footnote}}

\begin{abstract}

Re-identification (ReID) is a critical challenge in computer vision, predominantly studied in the context of pedestrians and vehicles. However, robust object-instance ReID, which has significant implications for tasks such as autonomous exploration, long-term perception, and scene understanding, remains underexplored. In this work, we address this gap by proposing a novel dual-path object-instance re-identification transformer architecture that integrates multimodal RGB and depth information. By leveraging depth data, we demonstrate improvements in ReID across scenes that are cluttered or have varying illumination conditions. Additionally, we develop a ReID-based localization framework that enables accurate camera localization and pose identification across different viewpoints. We validate our methods using two custom-built RGB-D datasets, as well as multiple sequences from the open-source TUM RGB-D datasets. Our approach demonstrates significant improvements in both object instance ReID (mAP of 75.18) and localization accuracy (success rate of 83\% on TUM-RGBD), highlighting the essential role of object ReID in advancing robotic perception. Our models, frameworks, and datasets have been made publicly available.
\footnote{\href{https://instance-based-loc-machine.github.io}{https://instance-based-loc-machine.github.io}}

\end{abstract}


\begin{teaserfigure}
    \centering
    \includegraphics[width=0.98\linewidth]{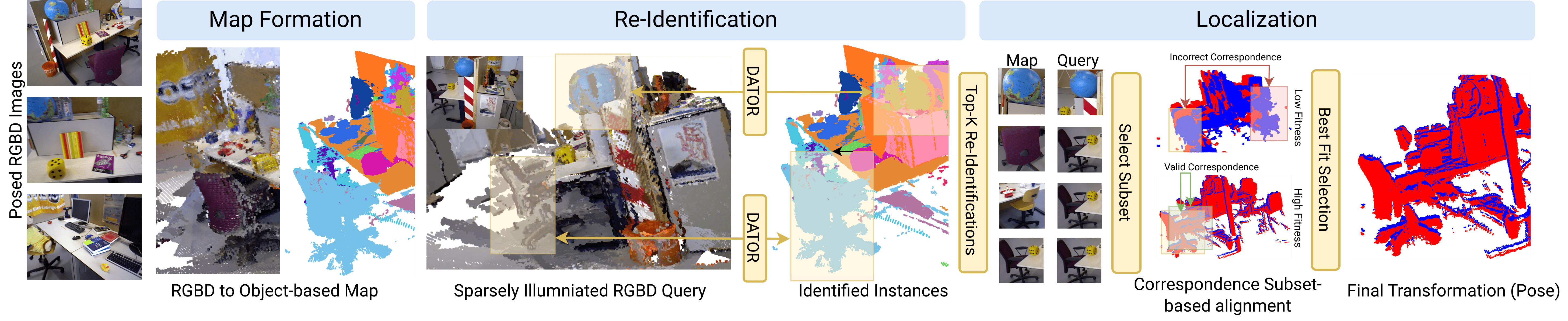}
    \caption{\emph{Overview of the Localization Framework.} \small{Our localization framework takes in posed-RGBD images and forms an object-based map consisting of instances and their descriptors. Given a query RGB-D image, we identify the objects within it and identify correspondences within our map using the ReID module (DATOR). The correspondences with the best fitness score are used to calculate the pose.}}
    \label{fig:loc_pipe}
\end{teaserfigure}

\received{11 February 2025}
\received[accepted]{18 April 2025}

\maketitle

\section{Introduction}

Objects in an environment can serve as important landmarks and offer significant cues for spatial awareness and orientation. They provide valuable information for understanding both an agent's general location and its precise orientation. However, the reliable re-identification of objects — formally known as the object-instance re-identification task — remains underexplored, particularly in the context of robotics.

Object-instance re-identification (ReID), often referred to simply as object ReID, is the task of reliably recognizing and matching identical instances of an object across different perspectives and environmental conditions. For example, in a warehouse setting, object ReID could be used to track the same piece of equipment across multiple camera views, even if the lighting or the equipment's position changes.  While extensive research has been conducted on ReID for specific categories such as people \cite{ye2021deep, tan2024harnessing, zhang2024view} and vehicles \cite{li2024day, zhu2020voc, amiri2024comprehensive}, which often leverage domain-specific features like gait patterns or vehicle parameters, the broader domain of object ReID presents unique challenges. Objects vary widely in structure, appearance, and type, lacking a common unifying feature. Foundational models such as DINOv2 \cite{dinov2} and vision-language models like CLIP \cite{clip} provide general classification into broad categories but fall short in re-identifying specific instances within these categories. Their ability to generalize to new scenes does not offer the fine-grained recognition needed for precise object-based applications.

\begin{figure}[t]
    \centering
    \includegraphics[width=0.98\linewidth]{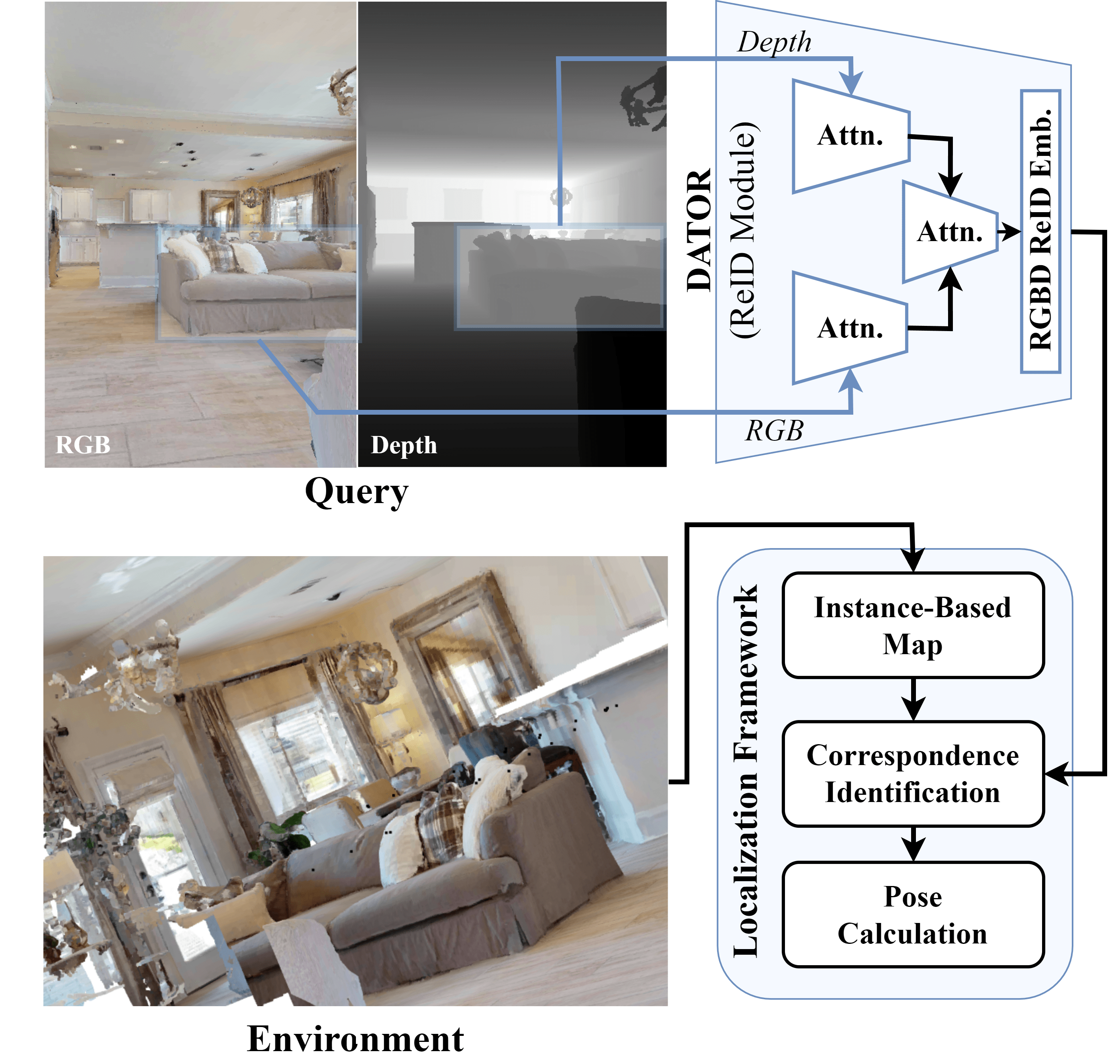}
    \caption{\emph{Overview.} \small{We propose a novel dual path transformer architecture, DATOR, combining cues from both RGB and depth modalities for effective object-instance ReID. Our localization framework generates an instance based map and uses our ReID model in conjunction with it to localize unseen views.}}
    \label{fig:teaser}
\end{figure}

In robotics, the ability to accurately re-identify objects can be widely utilized for various tasks. Global relocalization, in particular, is a critical application where accurate object ReID can significantly enhance performance. This task is especially challenging in environments with repetitive scenes or numerous objects and rooms, where both local and global registration difficulties are common. Traditional global relocalization approaches often rely on aligning entire point clouds \cite{elbaz20173d, besl1992method} or extensive collections of images \cite{zhang2021visual, ali2023mixvpr} to maximize available information. However, a significant portion of this information may be redundant or not informative for effective localization.


To tackle these challenges, we introduce a Dual Path Attention Transformer for Object Re-identification (DATOR), a deep object-ReID model that leverages RGB and depth sensors commonly used with mobile robots. DATOR employs a dual-path transformer architecture to significantly enhance its ReID capabilities across multiple views. This architecture extracts and refines features from both modalities, integrating them to produce a robust final embedding. By effectively combining these modalities, DATOR ensures high accuracy in object-ReID, maintaining performance across varying illumination conditions and diverse environmental settings.

Building on the fine-grained ReID capabilities of DATOR, we introduce an object-instance based global localization framework. (Fig. \ref{fig:loc_pipe}, \ref{fig:teaser}) This framework operates effectively in diverse indoor environments without requiring manual object annotation. Drawing inspiration from human navigation in familiar environments, our method constructs an instance-based map by mapping visible objects, aligning with principles similar to those discussed in \cite{zhang2023instaloc, matsuzaki2024clip}. We catalog and encode objects using our ReID model, preserving visual and structural information, while maintaining positional data through point clouds of individual objects. For localization, we process query RGB-D views to detect and match visible objects with those in our object map, optimizing alignment for accurate localization.

To validate our framework, we provide a real-world dataset from a large, object-rich laboratory environment with multiple instances per object class (e.g., tables, chairs), presenting a challenging ReID scenario. Additionally, we provide a real as well a synthetic indoor dataset with multiple objects of myriad classes for benchmarking global localization. We also benchmark our approach against sequences from the TUM \cite{sturm2012benchmark} RGB-D datasets. 

In conclusion, we make the following contributions:
\begin{itemize}
    \item A multimodal RGB-D object-instance ReID model (DATOR) achieving an mAP of $75.18$, higher than other SoTA models. 
    \item An object-instance ReID-based global localization framework, without manual annotation, for high accuracy in indoor environments, successfully localising $83.01\%$ of the time on dense, publicly available datasets.
    \item A comprehensive object-instance ReID dataset with multiple indoor object instances under varying lighting conditions.
    \item A real as well as a synthetic dataset for benchmarking global localization in complex indoor environments.
\end{itemize}

\section{Background}

\noindent\textbf{Re-Identification (ReID).} ReID is a computer vision task that focuses on the recognition and matching of specific objects or individuals across varying contexts and multiple camera views,  \emph{distinguishing it from general recognition and classification tasks} that typically involve identifying or categorizing objects \cite{ye2024transformer}. Extensive research has been done in person \cite{ye2021deep, tan2024harnessing, zhang2024view, LiPyramidal, dou2023identity, 8715446, 8053514} and vehicle ReID \cite{li2024day, zhu2020voc, amiri2024comprehensive, qian2022unstructured, zhao2021heterogeneous} to achieve near human performance on multiple datasets. Further, existing ReID methods also dive into the RGB-D and crossmodal domains \cite{9057668, li2023clipreid, 8053514, UDDIN2021100089} as well as ReID for animals and buildings \cite{jiao2024toward, Xue_2022_CVPR, adam2024seaturtleid2022}.

To the best of our knowledge, existing works that attempt to tackle object-instance ReID either use only one modality \cite{therien2024object, zhang2023instaloc, matsuzaki2024clip}, benchmark against person/vehicle ReID datasets rather than indoor objects \cite{zhu2022dual, lee2022negative, rao2021counterfactual}, focus on identifying evolving descriptions dependent on nearby objects \cite{keetha2022airobject} or operate on LiDAR scans \cite{pramatarov2024s}. 

\noindent\textbf{Foundational image models.} Foundational models refer to a category of models consisting of a very large number of parameters trained on a large and diverse variety of datasets for a particular task. Our framework uses three particular foundational models, Recognise Anything (RAM) \cite{zhang2023recognize}, Segment Anything (SAM) \cite{kirillov2023segany} and Grounding DINO \cite{gdino}. RAM is a captioning model that outputs a list of captions describing objects present in an input image. SAM is a general purpose segmentation model that can generate precise segmentation masks in challenging conditions given an image. DINOv2 \cite{dinov2} is all-purpose vision model trained at scale, and its variant Grounding DINO generates tight bounding boxes around an object.

\begin{figure*}[t!]
    \centering
    \includegraphics[width=0.9\linewidth]{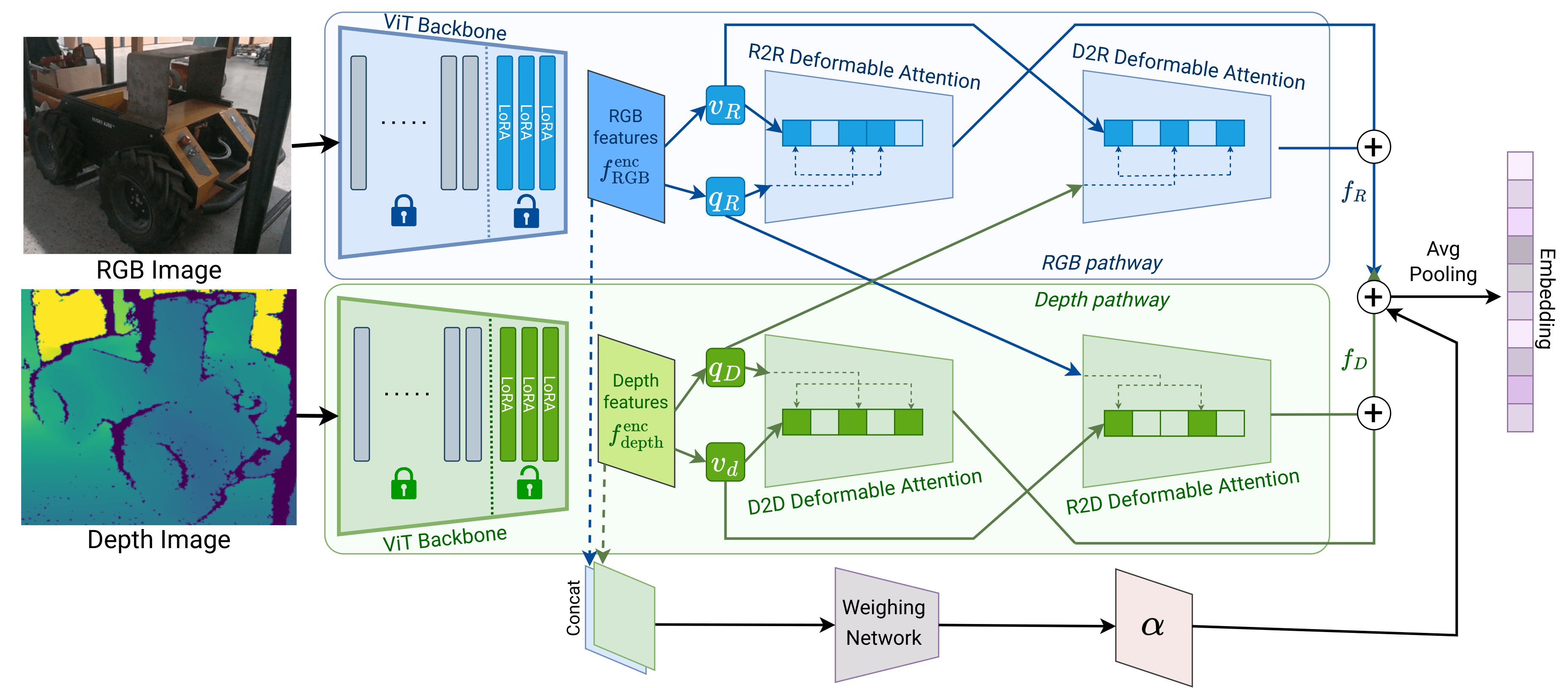}
    \caption{\emph{Proposed model \textit{DATOR}}: \small{The model can take paired RGB and depth images of an object, and utilize cues from both the modalities to give an embedding which can be used for object ReID.}}
    \label{fig:dual_path}
\end{figure*}

\noindent\textbf{Localization.} Localization is a well-studied problem in robotics with diverse solutions, including one-step methods \cite{panek2022meshloc, dong2022visual} and multi-step approaches \cite{wang2021continual}. Methods typically depend on scene as well as object recognition \cite{dong2022visual, peng2024slmsf, wen2024foundationpose, li2023magiccubepose, gard2024spvloc}. There are also approaches tailored for outdoor environments, which often deal with large-scale and diverse settings \cite{humenberger2022investigating, pramatarov2024s}, while indoor environments require more precise methods \cite{panek2022meshloc, yang2022scenesqueezer}. More novel approaches to localization have begun using neural fields \cite{aoki20243d, tang2024efficient}.

In contrast to the above methods, ReID-based localization focuses on matching individual object instances, disregarding the broader scene or surrounding objects. For example, \cite{matsuzaki2024clip} effectively demonstrates the capabilities of vision-language models (VLMs) like CLIP for object ReID, but struggles with scalability due to the use of generalized embeddings and the need for manual dataset annotation. 
On the other hand, \cite{zhang2023instaloc} is better suited for large-scale datasets but treats its environment as a collection of prior scans rather than a unified map, and relies solely on depth data, limiting its robustness in more complex scenarios.

\section{Approach}

\subsection{Dual Path Transformer Architecture}

\noindent\textbf{Network Architecture:} 
We propose a novel architecture for utilizing information from both the RGB and depth modalities (Fig. \ref{fig:dual_path}). The network has an RGB pathway and a depth pathway, taking an RGB and a depth image as input respectively. Within the network, information is exchanged between both the pathways, and finally features from both the pathways are combined to give a final embedding.

Specifically, the RGB and depth images are first input to their respective backbone networks to extract features $f_\text{depth}^\text{enc}, f_\text{RGB}^\text{enc}$ of size $(H \times W \times E)$, where $H, W$ represent the spatial dimensions and $E$ is the hidden dimension of the model. We use a ViT pre-trained on ImageNET \cite{imagenet} as the backbone for both the modalities. Additionally, we augment the last 2 encoder layers of both backbone ViTs with LoRA adapters \cite{hu2021lora}.


These features ($f_\text{depth}^\text{enc}, f_\text{RGB}^\text{enc}$) are further refined through specially designed attention modules. We use deformable attention~\cite{zhu2020deformable, xia2022vision} for our attention modules, an attention scheme that has been shown to address several limitations of using standard attention~\cite{vaswani2023attention}. Unlike in standard attention, where the final attention outputs are produced by a weighted sum of the \textit{entire} value feature map, in deformable attention, each query feature \textit{selects} a fixed number of positions ($K$) on the value feature map, and only those elements are used to calculate attention scores. For further details on deformable attention, we refer the reader to~\cite{zhu2020deformable}.

We linearly project RGB features $f^\text{enc}_\text{RGB}$ to compute queries $q_R$, and values $v_R$. Similarly, depth features $f^\text{enc}_\text{depth}$ are projected to obtain queries $q_D$ and values $v_D$. These query and value vectors are used for all subsequent attention calculations.

The RGB and the depth pathways undergo the following transformations: $$f_\text{R} = f_\text{RGB}^\text{enc} + \text{Attn}_\text{R2R}(q_R, v_R) + \text{Attn}_\text{D2R}(q_D, v_R)$$
$$f_\text{D} = f_\text{depth}^\text{enc} + \text{Attn}_\text{D2D}(q_D, v_D) + \text{Attn}_\text{R2D}(q_R, v_D)$$


where $f_\text{R}, f_\text{D}$ are of size $(H \times W \times E)$. Finally, we perform a learned weighted sum of features as 
$$f_\text{combined} = \alpha * f_\text{R} + (1 - \alpha) * f_\text{D}$$ with matrix $\alpha$ of size $(H \times W)$ effectively encoding the importance of each modality at each position on the feature map when making the final prediction. $\alpha$ is learnt by a CNN directly from the encoder feature representations $f^\text{enc}_\text{RGB}$ and $f^\text{enc}_\text{depth}$ (see Fig.~\ref{fig:dual_path} -- Weighing Network).  Finally, global average pooling is performed on this feature map to obtain the final embedding, a vector of dimension $E$. 





\noindent\textbf{Loss Function:}
Similar to earlier works~\cite{he2021transreid, ning2023occluded}, we use a cross entropy loss and a triplet loss, and the final loss function is given as  $\mathcal{L} = \mathcal{L}_\text{CE} + \mathcal{L}_\text{triplet}$.

\noindent\textbf{Modality Dropout:}
\label{subsec:dropout}
When training both the modalities, it is possible that one of the modality dominates the other. This may result in collapse for one of the two pathways. To prevent this, we employ \textit{modality dropout}~\cite{shi2022learning} during training, where for each training sample, we randomly zero-out features from one of the two modalities. The dropout is represented as:
$$
f_\text{RGB}^\text{enc}, f_\text{depth}^\text{enc} =
\begin{cases}
    f_\text{RGB}^\text{enc}, 0 & \text{with } p_\text{RGB} \\
    0, f_\text{depth}^\text{enc} & \text{with } p_\text{depth} \\
    f_\text{RGB}^\text{enc}, f_\text{depth}^\text{enc} & \text{with } 1 - p_\text{RGB} - p_\text{depth} \\
\end{cases}
$$

\subsection{Localization Framework}

Our localization framework (Fig. \ref{fig:loc_pipe}) aims to first build an object-instance based map of the environment, encoding information from each object it sees into separate reidentifiable units of information using embeddings from our ReID model. Then, when given an RGB-D image, it consults this map to find correspondences between visible objects and objects in it's memory. These correspondences are used to determine the RGB-D image's pose. 

\noindent\textbf{Memory formation.} Given a sequence of $n$ posed RGB-D images, $\{(I_i, D_i, t_i)\}_{i=1}^n$, with $I_i$, $D_i$, and $t_i$ representing the RGB image, depth image, and pose respectively, we build an object map $\mathcal{M} = \{\mathcal{O}_i\}_{i=1}^{n_\text{objects}}$, where each object $\mathcal{O}_i$ is stored as a tuple of its point cloud and embeddings (\emph{object info tuple}). RAM \cite{zhang2023recognize} outputs captions $c_i$, Grounding DINO \cite{gdino} generates bounding boxes $b_i$, and SAM \cite{kirillov2023segany} produces segmentation masks $m_i$. \textit{Filtering} removes captions that do not represent objects directly, like adjectives (``dark'', ``industrial'', ``wooden'') or descriptions of the whole scene itself (``living room'', ``workspace'').
\begin{align*}
\{c_1,\dots,c_p\} &= \operatorname{Filtering}(\operatorname{RAM}(I_i)) \\
\{b_1,\dots,b_j\} &= \operatorname{unique}(\operatorname{GDINO}(I_i, \{c_1,\dots,c_p\})) \\
m_k &= \operatorname{SAM}(I_i, b_k) \text{ for }  1 \leq k \leq j
\end{align*}
Using the RGB-D pair $(I_i, D_i)$, camera matrices $K$, world-frame transformation $T_i$, and segmentation mask $m_k$, we backproject each object and form its object info tuple:
\begin{align*}
\mathcal{P}_k^\text{global} &= f(I_i, D_i, m_k, K, T_i) \text{ for } 1 \leq k \leq j \\
e_k &= \operatorname{ReID}(I_i, D_i, b_k) \text{ for } 1 \leq k \leq j \\
\mathcal{O}_k &= (\mathcal{P}_k^\text{global}, [e_k]) \text{ for } 1 \leq k \leq j
\end{align*}

We repeat this for all RGB-D pairs in the sequence, generating the object memory from the object info tuples of detected objects in each image.


\noindent\textbf{Memory consolidation and post-processing}. 
In many cases, objects in a scene may be represented by multiple object information tuples after processing a sequence 
We can improve registration accuracy during localization by grouping together object tuples that belong to the same object to form more complete pointclouds and more representative sets of embeddings. To address this, we perform a postprocessing step that clusters object info tuples based on semantic similarity and spatial proximity. When combining $n$ tuples, their point clouds are combined, and all associated sets of embeddings are combined. We refer to this as \textit{grouping} tuples henceforth.
$$\mathcal{O}_i + \mathcal{O}_j := 
\{ \mathcal{P}_i \oplus \mathcal{P}_j , [e_{i_1},\dots,e_{i_m},e_{j_1},\dots,e_{j_n}] \}
$$




\begin{algorithm}
\caption{Object info tuple clustering}\label{alg:clustering}
\begin{algorithmic}

\State $\mathcal{M} \gets \{\mathcal{O}_1,\dots,\mathcal{O}_n\}$
\State $\epsilon_{\text{IoU}} \gets 0.25$
\State $\epsilon_{L_2} \gets 0.5$
\State Let ${\text{IoU}} \in \mathbb{R}^{n \times n}$
\ForAll{$(\mathcal{O}_i, \mathcal{O}_j)$ \textbf{in} $\mathcal{M} \times \mathcal{M}$}
    \State ${\text{IoU}}[i][j] = \operatorname{GetIoU}(\mathcal{O}_i, \mathcal{O}_j)$
\EndFor
%
\State ${\{\mathcal{L}_1, \dots, \mathcal{L}_n\}} \gets \ 
\operatorname{AggClustering}(\mathcal{M}, \text{IoU}, \epsilon_{\text{IoU}})$
\For {$k$ \textbf{in unique values in} $\{\mathcal{L}_1, \dots, \mathcal{L}_n\}$} 
\State $\mathcal{O}^\prime_k \gets \sum_{i=1}^{n} \mathcal{O}_i \text{ if } \mathcal{L}_i = k$
\EndFor
\State $\mathcal{M} \gets \{\mathcal{O}_1^\prime,\dots,\mathcal{O}_m^\prime\}$
\State Let ${\text{PairwiseDist}} \in \mathbb{R}^{m \times m}$
\ForAll{$(\mathcal{O}_i^\prime, \mathcal{O}_j^\prime)$ \textbf{in} $\mathcal{M} \times \mathcal{M}$}
    \State $\text{PairwiseDist}[i][j] = \operatorname{GetL2Distance}(\mathcal{O}_i^\prime, \mathcal{O}_j^\prime)$
\EndFor
\State ${\{\mathcal{L}_1^\prime, \dots, \mathcal{L}_m^\prime\}} \gets \ 
\operatorname{AggClustering}(\mathcal{M}, \text{PairwiseDist}, \epsilon_{L_2})$
\For {$k$ \textbf{in unique values in} $\{\mathcal{L}_1^\prime, \dots, \mathcal{L}_m^\prime\}$} 
\State $\{ d_1, \dots, d_a  \} \gets \ 
\operatorname{DBSCAN}(\{ \mathcal{O}^\prime_i \mid \mathcal{L}_i = k\})$
\For {$l$ \textbf{in unique values in} $\{ d_1, \dots, d_p  \}$}
\State $\mathcal{O}^{\prime\prime}_{k,l} \gets \sum_{i=0}^{m} \mathcal{O}^\prime_i \text{ if } \mathcal{L}_i = k , d_i = l$
\EndFor
\EndFor
\State $\mathcal{M} \gets \{\mathcal{O}_{1,1}^{\prime\prime},\dots,\mathcal{O}_{m,l}^{\prime\prime}\}$
\State \Return $\mathcal{M}$ 
\end{algorithmic}
\end{algorithm}



Clustering occurs in stages (Described in Algorithm \ref{alg:clustering}) to prevent errors from single-step grouping. Using just mean embedding-based clustering ignores positional data, potentially grouping distant objects together. To fix this, we apply DBSCAN \cite{Ester1996ADA} within each semantic cluster, ensuring object-instances are grouped based on both proximity in 3D space and embedding similarity. By performing clustering, each object has a more complete set of structural information, leading to more accurate point-feature generation and improved localization.


\noindent\textbf{Localization.} We begin by applying the RAM-Grounding DINO-SAM pipeline used during memory formation. By doing so, we obtain an object info tuple for each object. 



Since all partial point clouds are backprojected from the same RGB-D image, their relative positions remain unchanged after any rigid transformation. Therefore, the rigid transform aligning detected point clouds with those in object memory also localizes the RGB-D image in the global frame. We then seek the most accurate assignment between detected objects and those in memory using ReID embeddings, comparing assignments based on the $L2$ norm between ReID embeddings of detected objects from those in memory.

We score an assignment by the product of embedding distances between correspondences and select the $k$ lowest scores to avoid ICP failure from poor initialization. Using subsets of detected objects (containing at least 3 detections), we compute assignments to efficiently determine localization. Each assignment provides a transform that aligns detected objects with those in object memory. We combine point clouds from detected and memory objects into single \textit{detected} and \textit{memory} point clouds, respectively, and register them using RANSAC \cite{fischler1981random} followed by colored ICP. We enhance this process with custom features, including FPFH \cite{5152473} and one-hot encoded object indices, to prioritize matching between corresponding points. The quality of each pose is evaluated by the overlap between transformed detected point clouds and memory point clouds, with the best assignment being the one with the highest overlap.


\begin{table}[htbp]
\centering
\resizebox{\columnwidth}{!}{%
\begin{tabular}{|>{\centering\arraybackslash}p{0.2\linewidth}|
                >{\centering\arraybackslash}p{0.15\linewidth}|
                >{\centering\arraybackslash}p{0.15\linewidth}|
                >{\centering\arraybackslash}p{0.15\linewidth}|
                >{\centering\arraybackslash}p{0.15\linewidth}|}
\hline
\textbf{Dataset} & \textbf{Sequence Count} & \textbf{Mean seq. length} & \textbf{No. of classes} & \textbf{Mean instances per class} \\ \hline
\textit{DATOR-lab}   &  1&  1703&  19&  10\\ \hline
\textit{DATOR-synth} &  6&  3766&  9&  5\\ \hline
TUM                 &  5&  1331&  29&  2-3\\ \hline
\end{tabular}%
}

\caption{\emph{Localization Dataset Metrics}}
\label{tab:dataset_metrics}
\end{table}

\begin{table*}[h!]
\centering

\begin{tabular}{|c|c||c|c|c|}
\hline
\textbf{Dataset} & \textbf{ReID Backbone} & \textbf{Avg Errors (m, rad)} & \textbf{Median Errors (m, rad)} & \textbf{Success Rate (\%)} \\ \hline
\multirow{4}{6em}{TUM RGB-D} 
& \textbf{CLIP} & $1.83 / 0.9479$ & $1.60 / 1.121$ & $12.02$ \\ 
& \textbf{DINOv2} & $\underline{1.11} / \underline{0.629}$ & $\underline{0.85} / \underline{0.637}$ & $\underline{41.69}$ \\ 
& \textbf{ViT-b} & $1.32 / 0.720$ & $1.18 / 0.788$ & $24.70$ \\ 
& \textbf{DATOR} & \textbf{$\textbf{0.50} / \textbf{0.252}$} & \textbf{$\textbf{0.29} / \textbf{0.028}$} & \textbf{$\textbf{83.01}$} \\ \hline
\multirow{4}{6em}{DATOR-lab} 
& \textbf{CLIP} & $4.99 / 1.002$ & $4.74 / 1.211$ & $0.00$ \\ 
& \textbf{DINOv2} & $\underline{4.96} / 0.912$ & $4.69 / 1.01$ & $\underline{3.38}$ \\ 
&\textbf{ViT-b} & $5.28 / \underline{0.602}$ & $\underline{4.74} / \underline{0.322}$ & $0.00$ \\ 
& \textbf{DATOR} & \textbf{$\textbf{2.25} / \textbf{0.373}$} & \textbf{$\textbf{1.43} / \textbf{0.024}$} & \textbf{$\textbf{41.18}$} \\ \hline
\multirow{4}{6em}{DATOR-synth} 
& \textbf{CLIP} & $4.03 / \underline{0.327}$ & $\underline{2.26} / 0.00176$ & $\underline{33.33}$ \\ 
& \textbf{DINOv2} & $\underline{3.66}	/ \textbf{0.192}$ & $3.09 / 	\textbf{0.00099}$ & $15.38$ \\ 
&\textbf{ViT-b} & $5.11 / 0.341$ & $4.27 \underline{0.0017}$ & $12.82$ \\ 
&    \textbf{DATOR} & $\textbf{2.98}$ / $0.4552$ & $\textbf{0.93}$ / $0.0976$ & $\textbf{47.05}$ \\ \hline

\end{tabular}

\caption{\emph{Localization results across ReID techniques.} \small Avg. and Median Errors have been represented as (Translation Error/Rotation Error) in (m/rad) units. We consider a success as a pose prediction within $0.6$m of translation error and $0.3$ radians of rotation error. Our results show that DATOR outperforms other models by a significant margin across multiple datasets. Best results are in \textbf{bold} and second-best in \underline{underlined} text.}
\label{tab:loc_results}
\end{table*}
\vspace{-1cm}

\subsection{Dataset Generation}

We present \textit{DATOR-ReID}, a new, real-world, object re-identification dataset of RGB-D images designed to benchmark our object ReID model. Captured with an Intel RealSense \cite{keselman2017intelrealsensestereoscopicdepth} camera, it includes 8 types of objects, such as chairs and tables, with upto 5 distinct instances of each object type. We use approximately 60 images per instance under varying environmental conditions.




We also present two of our datasets, \textit{DATOR-lab} and \textit{DATOR-synth}, designed to evaluate our localization system. \textit{DATOR-lab} is a real-life localization dataset consisting collected on a P3DX robot in a large indoor laboratory. The RGB-D images were collected using an Intel RealSense camera and P3DX's wheel odometry. \textit{DATOR-synth}, is a set of sequences generated using the ProcTHOR \cite{deitke2022procthor} API in a multi-room environment with various items of furniture. Both datasets feature a large variety of objects and multiple instances of visually similar structures, resulting in a challenging localization benchmark. We note that \textit{DATOR-lab} and \textit{DATOR-ReID} were recorded in different locations. All of our data has been made publicly available
\footnote{\href{https://github.com/instance-based-loc/instance-based-loc}{https://github.com/instance-based-loc/instance-based-loc}}.

Table \ref{tab:dataset_metrics} collates metrics about the variety and number of objects in the each of the datasets we use in our localization experiments. 



\section{Experimental Setup}

\begin{table}[h]
\centering
\resizebox{\columnwidth}{!}{%
\begin{tabular}{|>{\centering\arraybackslash}p{0.55\linewidth}|
                >{\centering\arraybackslash}p{0.45\linewidth}|}
\hline
\textbf{Method} & \textbf{mAP Score} \\ \hline
        CLIP-ReiD (CNN) \cite{li2023clipreid} & 54.1\\ 
        TransReID \cite{he2021transreid} &  55.7\\ 
        LoGoViT \cite{10096126} &  60.2\\
        CLIP-ReID (ViT) \cite{li2023clipreid} & 61.2\\
        NFormer \cite{wang2022nformer} &  63.2\\ 
        PADE \cite{wang2024parallel} & \underline{66.0} \\ \hline
        
        \textbf{DATOR} (RGB Inference)&61.41\\ 
        \textbf{DATOR} (Depth Inference)& 64.11\\
        \textbf{DATOR} (Full)& \textbf{75.18}\\

        \hline
\end{tabular}%
}
\caption{\emph{ReID mAP Metrics.} \small{We benchmark DATOR against several ReID methods trained our \textit{DATOR-ReID} dataset. We demonstrate the effectiveness of using both modalities for object ReID. 
Best result is in \textbf{bold} and second-best is \underline{underlined}.}} 
\label{tab:reid_maps_table}
\end{table}

Our ReID module training as well as the localization pipeline requires only $12$ GB VRAM, allowing it to run on commercially available GPUs. Localization tests were conducted on an NVIDIA RTX A4000 GPU and an AMD Ryzen 9 7950X 16-Core Processor.

\noindent\textbf{Object ReID:} DATOR was trained for 240 epochs, using SGD optimizer and cosine LR scheduler, with an initial learning rate of 0.008. A batch size of 64 was used for all the experiments.


\noindent\textbf{Localization:}
We evaluate our localization pipeline on \textit{DATOR-lab}, 6  \textit{DATOR-synth} sequences, as well as 5 TUM RGB-D sequences, (namely \textit{fr\{\_desk,\_desk2,\_room\}}, \textit{freiburg2\_desk} and \\
\textit{freiburg3 \_long\_office\_household}). 

For \textit{DATOR-synth}, we create an object memory for each sequence by sampling RGB-D pairs every 30 frames. Additional unseen RGB-D pairs are sampled every 30 frames with a 15-frame offset. The only information available during localization is the generated object memory. Each frame is individually localized using the sequence's object memory. Sampling strategies for other datasets are detailed in our released implementation.

We compare each RGB-D pair's estimated pose with its corresponding ground truth pose, measuring 3D translation error (TE) in meters and rotation error (RE) in radians. An RGB-D pair is said to be correctly localized, if the TE and RE are below certain thresholds (see caption under table \ref{tab:loc_results}). We report the success rates for various baselines along with our method for comparison.

The implementations and annotation tools used for \cite{zhang2023instaloc} and \cite{matsuzaki2024clip} are not available. We release a re-implementation of \cite{matsuzaki2024clip} but observe suboptimal results using this approach (elaborated in section 5).

\section{Results}

\subsection*{Object ReID Results}

We benchmark DATOR against some popular vehicle-ReID and person-ReID baselines, the results are presented in table~\ref{tab:reid_maps_table}. We train all the RGB-only baselines on the RGB images in the \textit{DATOR-ReID} dataset. To evaluate the effectiveness of each individual pathway in DATOR, we also show RGB-only inference results, in which the depth features are zeroed (similar to~\hyperref[subsec:dropout]{modality dropout}) and depth-only inference results, wherein the RGB features are zeroed. Note that DATOR achieves competitive results even when using only one of the modalities, while at the same time achieving much better performance (last row in table~\ref{tab:reid_maps_table}) when both the modalities are used. This shows that there is effective exchange of information between both the pathways in the network.

\begin{figure}[hbt!]
    \centering
    \includegraphics[width=1.0\linewidth]{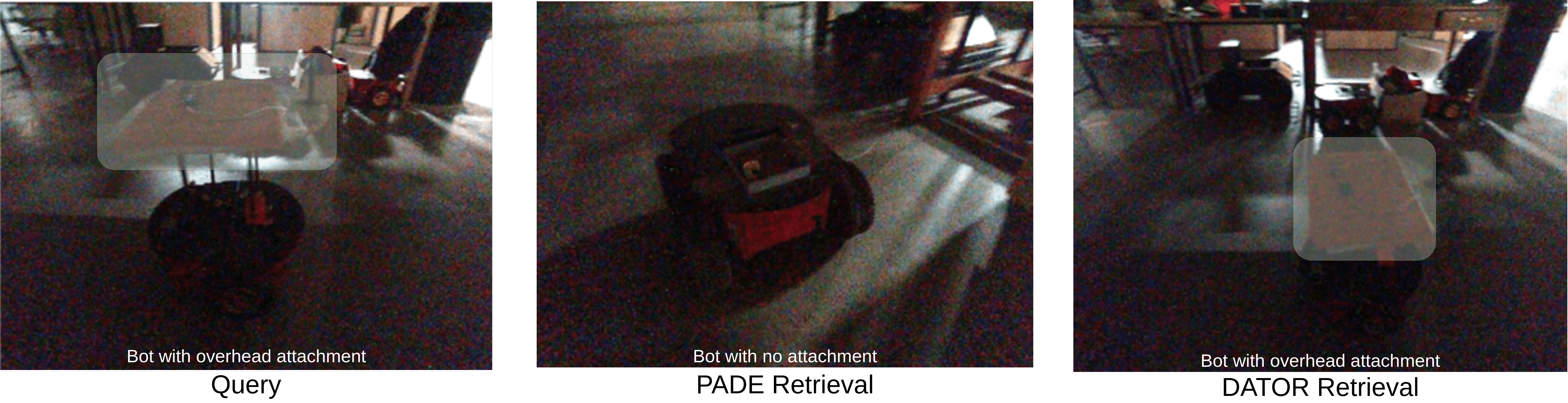}
    \caption{\emph{Qualitative Analysis of DATOR.} \small{Given a query in a low-illumination scene, DATOR reidentifies the robot instance successfully, while PADE, identifies it as a different robot that is missing an overhead attachment. This gain can be attributed to DATOR's use of depth information.}}
    \label{fig:dator-quality}
\end{figure}

Further, we show a qualitative result in Fig. \ref{fig:dator-quality} where DATOR re-identifies the correct instance in a low-illumination setting while the existing best performing method, PADE~\cite{wang2024parallel}, retrieves a similar but incorrect instance. 

\subsection*{Localization Results}

Table \ref{tab:loc_results} summarizes the effectiveness of each model in our localization architecture. We include both mean and median errors to illustrate the effect of high error misalignments as compared to successful alignments, that are near perfect predictions of pose. The large jump in success rate from other baselines to DATOR points to more successful instance ReID leading to more accurate registration. Our results demonstrate the challenging nature of our datasets and the potential improvements left to future works. A common trend observed across all models is the negative effect of large, flat, texture-less walls that interfere with ICP matching. For example, unlike the other TUM sequences, \textit{fr1\_room} dataset from TUM has large, textureless, flat objects that result in low success rates for all benchmarked methods.

In comparison, results shown in \cite{matsuzaki2024clip} are only for two TUM sequences (\textit{fr2\_desk}, \textit{fr3\_lo\_\_household}.) achieving close to only $80$\% while we achieve success rates of $88.5$\% and $100$\% respectively. This is despite of us employing stricter margins and also considering rotation error (See Fig. 4 in \cite{matsuzaki2024clip}).

\section{Conclusion}

We introduce an object-based localization framework that generalizes across diverse indoor environments without the need for manual annotations, marking a significant step forward in autonomous indoor navigation. Our approach demonstrates accurate and robust localization in both real-world and synthetic environments. Additionally, our ReID architecture achieves a high mean Average Precision (mAP) of $75.18$ on a challenging dataset with varying illumination, demonstrating its adaptability to real-world scenarios. The embeddings generated by our model allow for more accurate object-instance re-identification localization success rate of other large-scale image encoding models, localizing successfully in $83.01\%$ of the cases averaged across multiple sequences of TUM-RGBD. Moreover, we release a challenging, object-rich pair of real and synthetic relocalization datasets, as well as an object ReID dataset featuring varying illumination conditions.

Future work includes exploring expanding our framework to effectively recognise significant non-object landmarks, function in outdoor environments, enhancing its robustness to more extreme variations in lighting and occlusions and integrating it into mobile robotics pipelines.

\bibliographystyle{acm}
\bibliography{refs.bib}

\begin{thebibliography}{10}

\bibitem{adam2024seaturtleid2022}
{\sc Adam, L., {\v{C}}erm{\'a}k, V., Papafitsoros, K., and Picek, L.}
\newblock Seaturtleid2022: A long-span dataset for reliable sea turtle re-identification.
\newblock In {\em Proceedings of the IEEE/CVF Winter Conference on Applications of Computer Vision\/} (2024).

\bibitem{ali2023mixvpr}
{\sc Ali-Bey, A., Chaib-Draa, B., and Giguere, P.}
\newblock Mixvpr: Feature mixing for visual place recognition.
\newblock In {\em Proceedings of the IEEE/CVF winter conference on applications of computer vision\/} (2023).

\bibitem{amiri2024comprehensive}
{\sc Amiri, A., Kaya, A., and Keceli, A.~S.}
\newblock A comprehensive survey on deep-learning-based vehicle re-identification: Models, data sets and challenges.
\newblock {\em arXiv preprint arXiv:2401.10643\/} (2024).

\bibitem{aoki20243d}
{\sc Aoki, K., Koide, K., Oishi, S., Yokozuka, M., Banno, A., and Meguro, J.}
\newblock 3d-bbs: Global localization for 3d point cloud scan matching using branch-and-bound algorithm.
\newblock In {\em 2024 IEEE International Conference on Robotics and Automation (ICRA)\/} (2024), IEEE.

\bibitem{besl1992method}
{\sc Besl, P.~J., and McKay, N.~D.}
\newblock Method for registration of 3-d shapes.
\newblock In {\em Sensor fusion IV: control paradigms and data structures\/} (1992), vol.~1611, Spie.

\bibitem{deitke2022procthor}
{\sc Deitke, M., VanderBilt, E., Herrasti, A., Weihs, L., Salvador, J., Ehsani, K., Han, W., Kolve, E., Farhadi, A., Kembhavi, A., and Mottaghi, R.}
\newblock Procthor: Large-scale embodied ai using procedural generation, 2022.

\bibitem{dong2022visual}
{\sc Dong, S., Wang, S., Zhuang, Y., Kannala, J., Pollefeys, M., and Chen, B.}
\newblock Visual localization via few-shot scene region classification.
\newblock In {\em 2022 International Conference on 3D Vision (3DV)\/} (2022), IEEE, pp.~393--402.

\bibitem{dou2023identity}
{\sc Dou, Z., Wang, Z., Li, Y., and Wang, S.}
\newblock Identity-seeking self-supervised representation learning for generalizable person re-identification.
\newblock In {\em Proceedings of the IEEE/CVF international conference on computer vision\/} (2023).

\bibitem{elbaz20173d}
{\sc Elbaz, G., Avraham, T., and Fischer, A.}
\newblock 3d point cloud registration for localization using a deep neural network auto-encoder.
\newblock In {\em Proceedings of the IEEE conference on computer vision and pattern recognition\/} (2017), pp.~4631--4640.

\bibitem{Ester1996ADA}
{\sc Ester, M., Kriegel, H.-P., Sander, J., and Xu, X.}
\newblock A density-based algorithm for discovering clusters in large spatial databases with noise.
\newblock In {\em Knowledge Discovery and Data Mining\/} (1996).

\bibitem{fischler1981random}
{\sc Fischler, M.~A., and Bolles, R.~C.}
\newblock Random sample consensus: a paradigm for model fitting with applications to image analysis and automated cartography.
\newblock {\em Communications of the ACM 24}, 6 (1981).

\bibitem{gard2024spvloc}
{\sc Gard, N., Hilsmann, A., and Eisert, P.}
\newblock Spvloc: Semantic panoramic viewport matching for 6d camera localization in unseen environments.
\newblock {\em arXiv preprint arXiv:2404.10527\/} (2024).

\bibitem{he2021transreid}
{\sc He, S., Luo, H., Wang, P., Wang, F., Li, H., and Jiang, W.}
\newblock Transreid: Transformer-based object re-identification, 2021.

\bibitem{hu2021lora}
{\sc Hu, E.~J., Shen, Y., Wallis, P., Allen-Zhu, Z., Li, Y., Wang, S., Wang, L., and Chen, W.}
\newblock Lora: Low-rank adaptation of large language models, 2021.

\bibitem{humenberger2022investigating}
{\sc Humenberger, M., Cabon, Y., Pion, N., Weinzaepfel, P., Lee, D., Gu{\'e}rin, N., Sattler, T., and Csurka, G.}
\newblock Investigating the role of image retrieval for visual localization: An exhaustive benchmark.
\newblock {\em International Journal of Computer Vision 130}, 7 (2022), 1811--1836.

\bibitem{jiao2024toward}
{\sc Jiao, B., Liu, L., Gao, L., Wu, R., Lin, G., Wang, P., and Zhang, Y.}
\newblock Toward re-identifying any animal.
\newblock {\em Advances in Neural Information Processing Systems 36\/} (2024).

\bibitem{keetha2022airobject}
{\sc Keetha, N.~V., Wang, C., Qiu, Y., Xu, K., and Scherer, S.}
\newblock Airobject: A temporally evolving graph embedding for object identification.
\newblock In {\em Proceedings of the IEEE/CVF Conference on Computer Vision and Pattern Recognition\/} (2022).

\bibitem{keselman2017intelrealsensestereoscopicdepth}
{\sc Keselman, L., Woodfill, J.~I., Grunnet-Jepsen, A., and Bhowmik, A.}
\newblock Intel realsense stereoscopic depth cameras, 2017.

\bibitem{kirillov2023segany}
{\sc Kirillov, A., Mintun, E., Ravi, N., Mao, H., Rolland, C., Gustafson, L., Xiao, T., Whitehead, S., Berg, A.~C., Lo, W.-Y., Doll{\'a}r, P., and Girshick, R.}
\newblock Segment anything.
\newblock {\em arXiv:2304.02643\/} (2023).

\bibitem{lee2022negative}
{\sc Lee, H., Eum, S., and Kwon, H.}
\newblock Negative samples are at large: Leveraging hard-distance elastic loss for re-identification.
\newblock In {\em European Conference on Computer Vision\/} (2022), Springer.

\bibitem{8053514}
{\sc Lejbolle, A.~R., Nasrollahi, K., Krogh, B., and Moeslund, T.~B.}
\newblock Multimodal neural network for overhead person re-identification.
\newblock In {\em 2017 International Conference of the Biometrics Special Interest Group (BIOSIG)\/} (2017).

\bibitem{li2023magiccubepose}
{\sc Li, F., Gao, D., Huang, Q., Li, W., and Yang, Y.}
\newblock Magiccubepose, a more comprehensive 6d pose estimation network.
\newblock {\em Scientific Reports 13}, 1 (2023), 6923.

\bibitem{li2024day}
{\sc Li, H., Chen, J., Zheng, A., Wu, Y., and Luo, Y.}
\newblock Day-night cross-domain vehicle re-identification.
\newblock In {\em Proceedings of the IEEE/CVF Conference on Computer Vision and Pattern Recognition\/} (2024), pp.~12626--12635.

\bibitem{LiPyramidal}
{\sc Li, H., Ye, M., Wang, C., and Du, B.}
\newblock Pyramidal transformer with conv-patchify for person re-identification.

\bibitem{li2023clipreid}
{\sc Li, S., Sun, L., and Li, Q.}
\newblock Clip-reid: Exploiting vision-language model for image re-identification without concrete text labels, 2023.

\bibitem{gdino}
{\sc Liu, S., Zeng, Z., Ren, T., Li, F., Zhang, H., Yang, J., Jiang, Q., Li, C., Yang, J., Su, H., Zhu, J., and Zhang, L.}
\newblock Grounding dino: Marrying dino with grounded pre-training for open-set object detection, 2024.

\bibitem{9057668}
{\sc Martini, M., Paolanti, M., and Frontoni, E.}
\newblock Open-world person re-identification with rgbd camera in top-view configuration for retail applications.
\newblock {\em IEEE Access 8\/} (2020).

\bibitem{matsuzaki2024clip}
{\sc Matsuzaki, S., Sugino, T., Tanaka, K., Sha, Z., Nakaoka, S., Yoshizawa, S., and Shintani, K.}
\newblock Clip-loc: Multi-modal landmark association for global localization in object-based maps.
\newblock {\em International Conference on Robotics and Automation (ICRA)\/} (2024).

\bibitem{ning2023occluded}
{\sc Ning, E., Wang, C., Zhang, H., Ning, X., and Tiwari, P.}
\newblock Occluded person re-identification with deep learning: a survey and perspectives.
\newblock {\em Expert Systems with Applications\/} (2023), 122419.

\bibitem{dinov2}
{\sc Oquab, M., Darcet, T., Moutakanni, T., Vo, H., Szafraniec, M., Khalidov, V., Fernandez, P., Haziza, D., Massa, F., El-Nouby, A., Assran, M., Ballas, N., Galuba, W., Howes, R., Huang, P.-Y., Li, S.-W., Misra, I., Rabbat, M., Sharma, V., Synnaeve, G., Xu, H., Jegou, H., Mairal, J., Labatut, P., Joulin, A., and Bojanowski, P.}
\newblock Dinov2: Learning robust visual features without supervision, 2024.

\bibitem{panek2022meshloc}
{\sc Panek, V., Kukelova, Z., and Sattler, T.}
\newblock Meshloc: Mesh-based visual localization.
\newblock In {\em European Conference on Computer Vision\/} (2022), Springer.

\bibitem{peng2024slmsf}
{\sc Peng, Y., Zhai, Z., and Feng, M.}
\newblock Slmsf-net: A semantic localization and multi-scale fusion network for rgb-d salient object detection.
\newblock {\em Sensors 24}, 4 (2024), 1117.

\bibitem{10096126}
{\sc Phan, N., Huy, T.~D., Duong, S. T.~M., Hoang, N.~T., Tran, S., Hung, D.~H., Nguyen, C. D.~T., Bui, T., and Truong, S. Q.~H.}
\newblock Logovit: Local-global vision transformer for object re-identification.
\newblock In {\em ICASSP 2023 - 2023 IEEE International Conference on Acoustics, Speech and Signal Processing (ICASSP)\/} (2023).

\bibitem{pramatarov2024s}
{\sc Pramatarov, G., Gadd, M., Newman, P., and De~Martini, D.}
\newblock That's my point: Compact object-centric lidar pose estimation for large-scale outdoor localisation.
\newblock {\em arXiv preprint arXiv:2403.04755\/} (2024).

\bibitem{qian2022unstructured}
{\sc Qian, W., Luo, H., Peng, S., Wang, F., Chen, C., and Li, H.}
\newblock Unstructured feature decoupling for vehicle re-identification.
\newblock In {\em European Conference on Computer Vision\/} (2022), Springer, pp.~336--353.

\bibitem{clip}
{\sc Radford, A., Kim, J.~W., Hallacy, C., Ramesh, A., Goh, G., Agarwal, S., Sastry, G., Askell, A., Mishkin, P., Clark, J., Krueger, G., and Sutskever, I.}
\newblock Learning transferable visual models from natural language supervision, 2021.

\bibitem{rao2021counterfactual}
{\sc Rao, Y., Chen, G., Lu, J., and Zhou, J.}
\newblock Counterfactual attention learning for fine-grained visual categorization and re-identification.
\newblock In {\em Proceedings of the IEEE/CVF international conference on computer vision\/} (2021).

\bibitem{8715446}
{\sc Ren, L., Lu, J., Feng, J., and Zhou, J.}
\newblock Uniform and variational deep learning for rgb-d object recognition and person re-identification.
\newblock {\em IEEE Transactions on Image Processing 28}, 10 (2019), 4970--4983.

\bibitem{imagenet}
{\sc Russakovsky, O., Deng, J., Su, H., Krause, J., Satheesh, S., Ma, S., Huang, Z., Karpathy, A., Khosla, A., Bernstein, M., Berg, A.~C., and Fei-Fei, L.}
\newblock Imagenet large scale visual recognition challenge, 2015.

\bibitem{5152473}
{\sc Rusu, R.~B., Blodow, N., and Beetz, M.}
\newblock Fast point feature histograms (fpfh) for 3d registration.
\newblock In {\em 2009 IEEE International Conference on Robotics and Automation\/} (2009), pp.~3212--3217.

\bibitem{shi2022learning}
{\sc Shi, B., Hsu, W.-N., Lakhotia, K., and Mohamed, A.}
\newblock Learning audio-visual speech representation by masked multimodal cluster prediction.
\newblock {\em arXiv preprint arXiv:2201.02184\/} (2022).

\bibitem{sturm2012benchmark}
{\sc Sturm, J., Engelhard, N., Endres, F., Burgard, W., and Cremers, D.}
\newblock A benchmark for the evaluation of rgb-d slam systems.
\newblock In {\em 2012 IEEE/RSJ international conference on intelligent robots and systems\/} (2012), IEEE, pp.~573--580.

\bibitem{tan2024harnessing}
{\sc Tan, W., Ding, C., Jiang, J., Wang, F., Zhan, Y., and Tao, D.}
\newblock Harnessing the power of mllms for transferable text-to-image person reid.
\newblock In {\em Proceedings of the IEEE/CVF Conference on Computer Vision and Pattern Recognition\/} (2024).

\bibitem{tang2024efficient}
{\sc Tang, G., Jatavallabhula, K.~M., and Torralba, A.}
\newblock Efficient 3d instance mapping and localization with neural fields.
\newblock {\em arXiv preprint arXiv:2403.19797\/} (2024).

\bibitem{therien2024object}
{\sc Th{\'e}rien, B., Huang, C., Chow, A., and Czarnecki, K.}
\newblock Object re-identification from point clouds.
\newblock In {\em Proceedings of the IEEE/CVF Winter Conference on Applications of Computer Vision\/} (2024), pp.~8377--8388.

\bibitem{UDDIN2021100089}
{\sc Uddin, M.~K., Lam, A., Fukuda, H., Kobayashi, Y., and Kuno, Y.}
\newblock Fusion in dissimilarity space for rgb-d person re-identification.
\newblock {\em Array 12\/} (2021), 100089.

\bibitem{vaswani2023attention}
{\sc Vaswani, A., Shazeer, N., Parmar, N., Uszkoreit, J., Jones, L., Gomez, A.~N., Kaiser, L., and Polosukhin, I.}
\newblock Attention is all you need, 2023.

\bibitem{wang2022nformer}
{\sc Wang, H., Shen, J., Liu, Y., Gao, Y., and Gavves, E.}
\newblock Nformer: Robust person re-identification with neighbor transformer, 2022.

\bibitem{wang2021continual}
{\sc Wang, S., Laskar, Z., Melekhov, I., Li, X., and Kannala, J.}
\newblock Continual learning for image-based camera localization.
\newblock In {\em Proceedings of the IEEE/CVF International Conference on Computer Vision\/} (2021), pp.~3252--3262.

\bibitem{wang2024parallel}
{\sc Wang, Z., Huang, H., Zheng, A., Li, C., and He, R.}
\newblock Parallel augmentation and dual enhancement for occluded person re-identification, 2024.

\bibitem{wen2024foundationpose}
{\sc Wen, B., Yang, W., Kautz, J., and Birchfield, S.}
\newblock Foundationpose: Unified 6d pose estimation and tracking of novel objects.
\newblock In {\em Proceedings of the IEEE/CVF Conference on Computer Vision and Pattern Recognition\/} (2024), pp.~17868--17879.

\bibitem{xia2022vision}
{\sc Xia, Z., Pan, X., Song, S., Li, L.~E., and Huang, G.}
\newblock Vision transformer with deformable attention.
\newblock In {\em IEEE/CVF CVPR\/} (2022).

\bibitem{Xue_2022_CVPR}
{\sc Xue, F., Budvytis, I., Reino, D.~O., and Cipolla, R.}
\newblock Efficient large-scale localization by global instance recognition.
\newblock In {\em Proceedings of the IEEE/CVF Conference on Computer Vision and Pattern Recognition (CVPR)\/} (6 2022).

\bibitem{yang2022scenesqueezer}
{\sc Yang, L., Shrestha, R., Li, W., Liu, S., Zhang, G., Cui, Z., and Tan, P.}
\newblock Scenesqueezer: Learning to compress scene for camera relocalization.
\newblock In {\em Proceedings of the IEEE/CVF conference on computer vision and pattern recognition\/} (2022).

\bibitem{ye2024transformer}
{\sc Ye, M., Chen, S., Li, C., Zheng, W.-S., Crandall, D., and Du, B.}
\newblock Transformer for object re-identification: A survey, 2024.

\bibitem{ye2021deep}
{\sc Ye, M., Shen, J., Lin, G., Xiang, T., Shao, L., and Hoi, S.~C.}
\newblock Deep learning for person re-identification: A survey and outlook.
\newblock {\em IEEE transactions on pattern analysis and machine intelligence 44}, 6 (2021), 2872--2893.

\bibitem{zhang2023instaloc}
{\sc Zhang, L., Digumarti, T., Tinchev, G., and Fallon, M.}
\newblock Instaloc: One-shot global lidar localisation in indoor environments through instance learning.
\newblock {\em Robotics: Science and Systems\/} (2023).

\bibitem{zhang2024view}
{\sc Zhang, Q., Wang, L., Patel, V.~M., Xie, X., and Lai, J.}
\newblock View-decoupled transformer for person re-identification under aerial-ground camera network.
\newblock In {\em Proceedings of the IEEE/CVF Conference on Computer Vision and Pattern Recognition\/} (2024), pp.~22000--22009.

\bibitem{zhang2021visual}
{\sc Zhang, X., Wang, L., and Su, Y.}
\newblock Visual place recognition: A survey from deep learning perspective.
\newblock {\em Pattern Recognition 113\/} (2021), 107760.

\bibitem{zhang2023recognize}
{\sc Zhang, Y., Huang, X., Ma, J., Li, Z., Luo, Z., Xie, Y., Qin, Y., Luo, T., Li, Y., Liu, S., et~al.}
\newblock Recognize anything: A strong image tagging model.
\newblock {\em arXiv preprint arXiv:2306.03514\/} (2023).

\bibitem{zhao2021heterogeneous}
{\sc Zhao, J., Zhao, Y., Li, J., Yan, K., and Tian, Y.}
\newblock Heterogeneous relational complement for vehicle re-identification.
\newblock In {\em Proceedings of the IEEE/CVF International Conference on Computer Vision\/} (2021), pp.~205--214.

\bibitem{zhu2022dual}
{\sc Zhu, H., Ke, W., Li, D., Liu, J., Tian, L., and Shan, Y.}
\newblock Dual cross-attention learning for fine-grained visual categorization and object re-identification, 2022.

\bibitem{zhu2020voc}
{\sc Zhu, X., Luo, Z., Fu, P., and Ji, X.}
\newblock Voc-reid: Vehicle re-identification based on vehicle-orientation-camera.
\newblock In {\em Proceedings of the IEEE/CVF Conference on Computer Vision and Pattern Recognition Workshops\/} (2020), pp.~602--603.

\bibitem{zhu2020deformable}
{\sc Zhu, X., Su, W., Lu, L., Li, B., Wang, X., and Dai, J.}
\newblock Deformable detr: Deformable transformers for end-to-end object detection.
\newblock {\em arXiv preprint arXiv:2010.04159\/} (2020).

\end{thebibliography}

\appendix

\end{document}